# FSGNet: A Frequency-Aware and Semantic Guidance Network for Infrared Small Target Detection

Yingmei Zhang, Wangtao Bao, Yong Yang, *Senior Member, IEEE*, Weiguo Wan *Member, IEEE*, Qin Xiao, and Xueting Zou

*Abstract*—Infrared small target detection (IRSTD) aims to identify and distinguish small targets from complex backgrounds. Leveraging the powerful multi-scale feature fusion capability of the U-Net architecture, IRSTD has achieved significant progress. However, U-Net suffers from semantic degradation when transferring high-level features from deep to shallow layers, limiting the precise localization of small targets. To address this issue, this paper proposes FSGNet, a lightweight and effective detection framework incorporating frequency-aware and semantic guidance mechanisms. Specifically, a multi-directional interactive attention module is proposed throughout the encoder to capture fine-grained and directional features, enhancing the network's sensitivity to small, low-contrast targets. To suppress background interference propagated through skip connections, a multi-scale frequency-aware module leverages Fast Fourier transform to filter out target-similar clutter while preserving salient target structures. At the deepest layer, a global pooling module captures high-level semantic information, which is subsequently upsampled and propagated to each decoder stage through the global semantic guidance flows, ensuring semantic consistency and precise localization across scales. Extensive experiments on four public IRSTD datasets demonstrate that FSGNet achieves superior detection performance and maintains high efficiency, highlighting its practical applicability and robustness. The codes will be released on https://github.com/Wangtao-Bao/FSGNet.

## I. INTRODUCTION

Infrared small target detection (IRSTD) plays a key role in various real-world scenarios, including environmental monitoring, security surveillance, and autonomous navigation systems [1], [2], [3]. Infrared detectors can identify targets at long distances (usually up to hundreds of kilometers), causing them to appear faint and difficult to distinguish in infrared imagery. These small targets typically lack distinct texture, shape, and color features, and are easily interfered by complex background elements and inherent sensor noise during the imaging process, making them susceptible to occlusion and confusion with background clutter. Effectively detecting and segmenting infrared small targets while addressing background complexity and noise interference remains a critical challenge in the field of object detection.

Over the past few decades, researchers have proposed various traditional IRSTD methods based on hand-crafted feature, including background estimation [4], local contrast enhancement [5], and low-rank decomposition techniques [6]. However, these methods heavily rely on prior knowledge and often struggle to capture high-level semantic representations of the image content.

With the advancement of deep learning, IRSTD has witnessed remarkable progress, primarily driven by the adoption of convolutional neural networks (CNNs) for feature extraction. In particular, U-Net-based architectures are extensively used due to their encoder–decoder structure and ability to combine multi-level features [7]. ACM [8] pioneered the first U-Net-based IRSTD method by introducing a cross-layer feature integration module that employs an asymmetric top-down and bottom-up structure to merge high-level semantic information with low-level details. DNANet [9] further proposed a dense nested interaction module to enhance progressive interaction fusion across hierarchical features. UIUNet [10] extended this concept by embedding multiple smaller U-Net structures within a larger framework to capture local target contrast more effectively, employing an interactive cross-attention mechanism for feature refinement. SCTransNet [11] integrated multiple spatial/channel cross transformer blocks into the skip connections of the original U-Net structure to improve the semantic distinction between foreground targets and cluttered backgrounds.

While these methods improve feature fusion and representation, they largely inherit the structural characteristics of the U-Net decoder, which may limit their ability to preserve semantic information during reconstruction. A critical limitation of the U-Net structure lies in the upsampling process of the decoder, where the high-level

This work was supported in part by the National Natural Science Foundation of China under Grant 62072218 and 62261025, in part by Gan-Po Talents Support Program—Urgently Needed Overseas Talent Program under Grant 20242BCE50059, by Early-Career Young Science and Technology Talent Training Project of Jiangxi Province under Grant 20244BCE52075 and 20244BCE52088, in part by Graduate Innovation Special Foundation Project of Jiangxi Province under Grant YC2024-S375. (Corresponding authors: Yingmei Zhang and Yong Yang)

Y. Zhang, W. Bao, W. Wan, X. Zou are with the School of Software and Internet of Things Engineering, Jiangxi University of Finance and Economics, Nanchang 330032, China (e-mail: zhangyingmei@jxufe.edu.cn; baowangtao29@gmail.com;wanweiguo@jxufe.edu.cn;zouxueting0113@126.com). Y. Zhang is also with the postdoctoral researcher, CGN Begood Technology co., Ltd, Nanchang 330032, China.

Y. Yang is with the School of Computer Science and Technology, Tiangong University, Tianjin 300387, China (e-mail:greatyangy@126.com).

Q. Xiao is with the School of Information Management and Mathematics, Jiangxi University of Finance and Economics, Nanchang 330032, China (e-mail: xiaoqin@jxufe.edu.cn).

semantic representations learned in the deepest layers tend to be diluted by low-level features propagated from earlier stages. While previous studies have attempted to mitigate this problem through enhanced skip connections or multi-scale fusion mechanisms, the output of the decoder often still lacks sufficient semantic richness to enable precise localization of small infrared targets, particularly in complex backgrounds.

To address this issue, inspired by [12], this paper proposes FSGNet, a lightweight and effective framework that integrates three functionally complementary modules: a multi-directional interactive attention module (MIAM), a multi-scale frequency-aware module (MFM), and a global pooling module (GPM). To begin with, inspired by [13], [14], [15], MIAM is proposed throughout the encoder to enhance local feature representation. Leveraging a novel pinwheel-shaped convolution (PConv), it captures directional and structural information from multiple orientations, improving the detection ability in cluttered backgrounds. To suppress background interference introduced via skip connections, MFM is designed before decoder fusion. By projecting features into the frequency domain using Fast Fourier Transform (FFT), it filters out target-similar clutter and improves target saliency. At the deepest layer of the network, GPM aggregates global semantic context, which is then upsampled and transmitted to decoder layers through semantic guidance flows, enhancing spatial localization across scales. Experimental results on four public datasets show that the proposed method outperforms several state-of-the-art (SOTA) methods.

The main contributions of this paper are as follows.

1) FSGNet, a novel lightweight framework for infrared small target detection, is proposed by unifying frequency-aware and semantic guidance design to enhance localization accuracy and suppress background interference.

2) The MIAM is constructed to enhance local feature representation by capturing structural and directional patterns from multiple orientations, thereby improving target saliency and robustness in cluttered, low-contrast infrared scenes.

3) The MFM is developed to suppress background interference in skip connections by operating in the frequency domain via Fast Fourier Transform, effectively filtering target-like clutter and enhancing feature reliability in complex infrared scenes.

4) Extensive experiments on four public IRSTD benchmarks demonstrate that FSGNet outperforms existing SOTA methods in terms of detection accuracy, generalization capability, and computational efficiency.

The reminder of this paper is structured as follows. Section II gives a brief introduction of related work. Section III details the architecture of FSGNet along with its three core modules: MIAM, MFM, and GPM. Section IV discusses the experimental results, and Section V concludes the paper.

## II. RELATED WORK

### A. Infrared Small Target Detection

IRSTD methods are generally divided into two categories: traditional and deep learning-based methods. Traditional IRSTD methods primarily rely on handcrafted feature engineering, emphasizing differences in brightness, contrast, and intensity between the target and the background. These methods are typically grouped into three major types. *1) Filter-based methods* aim to suppress background clutter leveraging various filtering techniques, such as bilateral filtering [16], Top-Hat filtering [17], [18], and maximum mean/median filtering [19]. *2) Human Visual System-based methods* detect small targets by analyzing local contrast differences between targets and their surrounding background. For instance, local contrast measurement (LCM) [20] and multiscale patch contrast measurement [21]. *3) Low-rank-based methods* [22], [23], [24] formulate small target detection as a convex optimization problem for recovering low-rank sparse matrix, as exemplified by local [22] and non-local [23] low-rank models. These methods are particularly effective in scenarios characterized by low signal-to-clutter ratio and complex backgrounds. However, despite ongoing development, traditional methods still face significant limitations in practice. Their reliance on handcrafted features and fixed hyperparameters reduces adaptability, particularly in real-world scenarios with dynamic and cluttered backgrounds.

Deep learning-based methods have significantly advanced IRSTD by leveraging neural networks to capture complex mappings between real-world scenes and infrared imagery. MDvsFA-cGAN [25] introduced constraints on false alarms and missed detections during training, however, it failed to fully utilize local target features. To address this, ALCNet [26] incorporates an attention module to enhance local feature representation. ISNet [27] proposed a Taylor differential operator to strengthen edge information. AGPCNet [28] developed an attention-guided semantic module that improves semantic encoding and feature integration for small targets. BAFE-Net [29] proposed a background-aware feature exchange network for clustered IRSTD, enabling effective deep features extraction from both targets and background. IRSAM [30] redesigned the general visual segmentation framework SAM to better suit the characteristics of IRSTD and achieved SOTA detection performance in most typical scenarios. As conventional CNN kernels are often insensitive to subtle infrared features, L$^2$SKNet [31] replaces standard cross-layer fusion with a learnable local saliency kernel module to better highlight target-relevant information. Moreover, the emerging Mamba [32] architecture has garnered increasing attention due to its strong capability to model long-range dependencies, prompting initial exploration of its application to IRSTD tasks. MiM-ISTD [33] was the first to apply Mamba to IRSTD, introducing a hierarchical Mamba-in-Mamba structure to model both global and local dependencies efficiently by treating image patches as visual sentences and words. IRMamba [34] further enhanced target detail representation by utilizing the pixel differences between the scanning position and its surrounding neighborhood, incorporating this information into the state equation of the state space model.

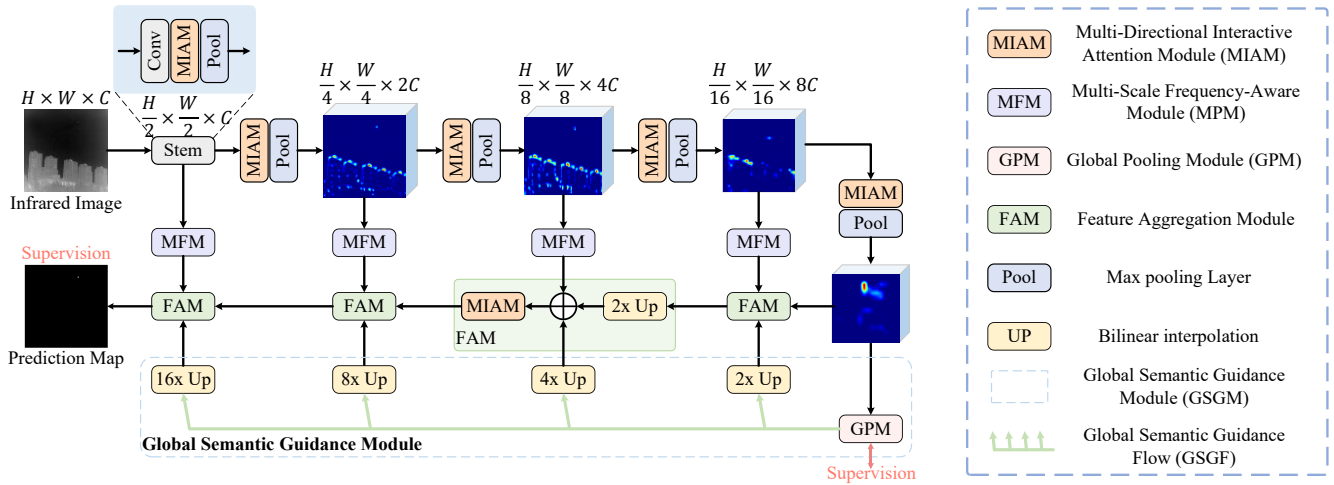

Fig. 1. Overall architecture of the proposed FSGNet.

## B. Image Processing in Frequency-domain

Beyond spatial-domain modeling, frequency-domain analysis has emerged as a powerful tool for image enhancement and feature discrimination. Classical frequency-domain techniques rely on applying the FFT to convert spatial data into the frequency domain, enabling specific operations such as filtering or enhancement, followed by reconstruction through inverse FFT.

In recent years, frequency-domain modeling has shown increasing potential in computer vision tasks [35] [36] [37], including object detection and semantic segmentation. This progress has also inspired developments in the IRSTD domain. FDDBA-Net [38] proposed a frequency-domain decoupling mechanism to isolate target features from background interference. FDA-IRSTD [39] introduced a block-level FFT framework to separate high- and low-frequency components, enhancing target saliency. HLSR-Net [40] further refined semantic reconstruction using high- and low-frequency cues, while HDNet [41] designed a dynamic high-pass filter to preserve discriminative high-frequency signals.

These advances collectively highlight the robustness and efficacy of frequency-domain learning. Motivated by this, the proposed FSGNet incorporates a dedicated multi-scale frequency-aware module to refine skip-connections and improve detection reliability by leveraging frequency-domain priors. This integration complements the spatial-domain semantic guidance design, forming a unified architecture optimized for accurate and robust small target detection.

## III. EXPERIMENTS

### A. Overall Architecture

The proposed network builds on the U-Net architecture, which has been widely used in IRSTD due to its effectiveness in multi-scale feature fusion [42], [43], [44], [45]. As illustrated in Fig. 1, three synergistic modules are embedded into the U-Net framework, forming a coordinated mechanism to improve the network's robustness in detecting small, low-contrast targets amidst cluttered backgrounds.

In this framework, a multi-directional interactive attention module (MIAM) is proposed as the basic feature extraction module to enhance the analysis of the underlying features of infrared small targets. At the same time, skip connections—though beneficial for feature reuse—can introduce background noise into the decoder. To alleviate this, a multi-scale frequency-aware module (MFM) is designed along each skip pathway. By transforming features into the frequency domain, MFM suppresses target-like background interference and emphasizes target-relevant responses. At the bottom of the U-Net, a global pooling module (GPM) is constructed to capture high-level semantic information. This global context is upsampled and injected into each decoder stage via global semantic guidance flows (GSGFs), ensuring that semantic cues are consistently reinforced throughout the decoding process. This design helps compensate for missing semantic information in shallower layers, thereby improving small target localization. The structure and functionality of each module will be described in the following sections.

### B. Multi-directional Interactive Attention Module

Previous IRSTD methods [8], [11], [26] often employed classical residual structures [13] for feature extraction. However, traditional CNNs are limited by fixed-size convolutional kernels, resulting in a constrained receptive field and limited capacity to capture long-range dependencies. Under such conditions, relying solely on local convolutions hinders the network's ability to model global contextual information, ultimately compromising detection performance.

To alleviate this issue, a MIAM is introduced, as illustrated in Fig. 2. MIAM adopts pinwheel-shaped convolution (PConv) [14] as the core feature extraction unit, and integrates batch normalization, ReLU activation, and a convolutional block attention module (CBAM) [15] to enhance feature representation and improve sensitivity to small, weak targets.

#### 1) Basic Feature Extraction

Unlike traditional convolution operations that implicitly rely on the Gaussian spatial distribution of infrared small targets, the PConv introduces asymmetric padding to construct direction-specific kernels. By applying distinct horizontal and vertical convolutions to different regions of the image, PConv enables the network to capture features along multiple orientations, thereby improving its adaptability to the varied morphology of infrared small targets.

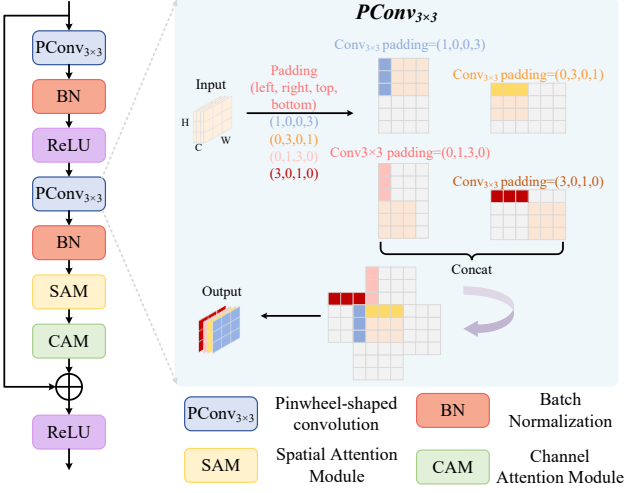

Fig. 2. Detailed structure of the proposed MIAM.

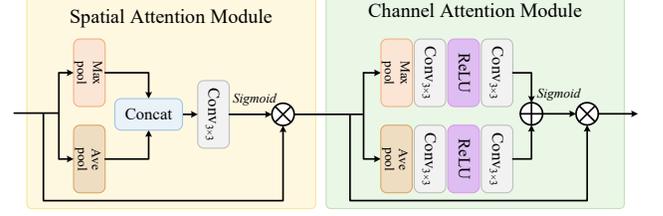

Fig. 3. Detailed structure of the spatial attention module and channel attention module.

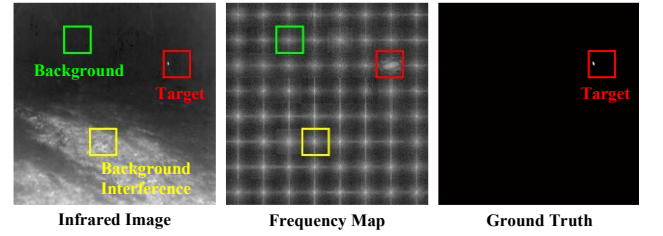

Fig. 4. The frequency map of each image patch illustrates that the green, yellow, and red bounding boxes denote the background, high-similarity background interference, and small target, respectively.

Given the input tensor $X^{(h,w,c)}$, where $h$, $w$, $c$ represent the height, width, and number of channels respectively, the first layer of PConv performs parallel convolution as follows:

$$X_1^{(h,w,c)} = X_{p(1,0,0,3)}^{(h,w,c)} \odot C_1^{(3,1,c)}, \quad (1)$$

$$X_2^{(h,w,c)} = X_{p(0,3,0,1)}^{(h,w,c)} \odot C_2^{(1,3,c)}, \quad (2)$$

$$X_3^{(h,w,c)} = X_{p(0,1,3,0)}^{(h,w,c)} \odot C_3^{(3,1,c)}, \quad (3)$$

$$X_4^{(h,w,c)} = X_{p(3,0,1,0)}^{(h,w,c)} \odot C_4^{(1,3,c)}. \quad (4)$$

where $\odot$ is the convolution operator, $C_1^{(1,3,c)}$ is a 1×3 convolution kernel with an output channel of $c$. The padding parameters $p(1,0,0,3)$ denote the number of padding pixels in the left, right, top, and bottom directions, respectively. The results of the first layer of parallel convolution are then concatenated in the channel dimension, and the output is calculated as:

$$X'^{(h,w,4c)} = Concat(X_i^{(h,w,c)}) \ (i = 1,2,3,4), \quad (5)$$

Finally, the concatenated tensor is normalized by the convolution kernel $C^{(2,2,c2)}$ without padding, where $c2$ is the number of channels of the final output feature map of the PConv module. The final output tensor $Y^{(h_2,w_2,c_2)}$ is calculated as:

$$Y^{(h_2,w_2,c_2)} = X'^{(h,w,4c)} \odot C^{(2,2,c2)} \quad (6)$$

Compared with traditional convolutional layers, this model uses asymmetric convolution with multi-directional padding to simultaneously increase the receptive field from multiple directions, which is more in line with the distribution of small targets and can extract more detailed information, thereby improving the detection capability of small targets.

*2) Convolutional Block Attention Module*

Following the multi-directional interactive feature extraction, CBAM, as illustrated in Fig. 3, is introduced to enhance the feature representations in both spatial and channel dimensions, thereby enabling the network to focus on salient target regions. Given the input feature $F$, the computation of spatial and channel attention is formulated as follows:

$$F_{CBAM} = C_{att}(S_{att}(F)), \quad (7)$$

where $F_{CBAM}$ denotes the output of two mechanisms, $S_{att}$ and $C_{att}$ represent the spatial attention and channel attention modules, respectively.

Specifically, the spatial attention module (SAM) improves localization by generating attention maps that capture spatial dependencies within the input feature map $F_{in}$. The computation process is defined as follows:

$$S_{att}(F_{in}) = \sigma\big(Conv_{7\times7}[\mathcal{P}_{max}(F_{in}), \mathcal{P}_{avg}(F_{in})]\big) \otimes F_{in}, \quad (8)$$

where $\mathcal{P}_{max}$ and $\mathcal{P}_{avg}$ denote max-pooling and average-pooling operations, respectively. $Conv_{7\times7}$ is a $7 \times 7$ convolutional layer for feature aggregation. $\sigma$ represents the sigmoid activation function. $\otimes$ is the element-wise multiplication, modulating the original features with the computed attention map.

The channel attention module (CAM) adaptively assigns weights to feature channels, enhancing informative representations while suppressing less relevant responses. Its mechanism is mathematically defined as follows:

$$C_{att}(F_{in}) = \sigma\begin{pmatrix} MLP(\mathcal{P}_{max}(F_{in})) \\ \oplus MLP(\mathcal{P}_{avg}(F_{in})) \end{pmatrix} \otimes F_{in}, \quad (9)$$

where $MLP(\cdot)$ is a multi-layer perceptron with one hidden layer, $\oplus$ is an element-wise addition operation.

*C. Multi-scale Frequency-aware Module*

Infrared images often contain background interference that closely resembles true small targets, leading to a high risk of false alarms. These false alarms may propagate through the skip connections and degrade detection performance. Frequency domain features, known for their robustness to local noise and interference [2], help distinguish targets from background clutter by capturing distinct spectral patterns, as illustrated in Fig. 4. The spectral distribution of a small target region (red box) differs significantly from that of a visually similar background region (yellow box) in the frequency domain. This distinction enables frequency features to more effectively separate targets from interference, thereby enhancing detection accuracy. Moreover, spatial and frequency domain features are inherently complementary: while the frequency domain captures global structural and periodic patterns, spatial domain feature extraction benefits from a larger receptive field [45], allowing it to encode intuitive image characteristics such as edges, textures, and shapes. The integration of these two domains facilitates the modeling of multi-scale contextual information, thereby enhancing overall feature representation. Motivated by these observations, a MFM is designed to achieve local-global feature fusion with a spatial-frequency cascade structure, as illustrated in Fig. 5.

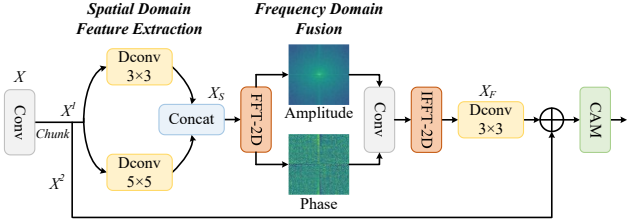

Fig. 5. Detailed structure of the proposed MFM.

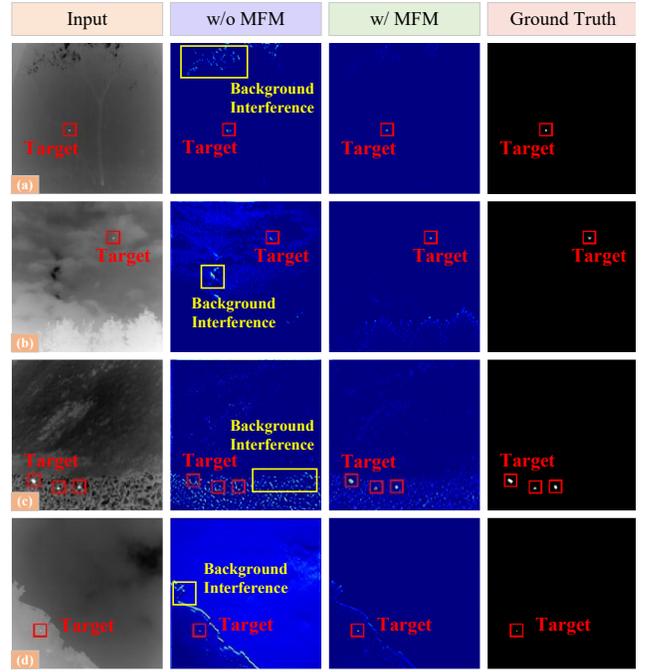

Fig. 6. Visualization results of feature maps with and without MPM. Each column shows the infrared image, the feature maps with and w/o, and the corresponding ground truth.

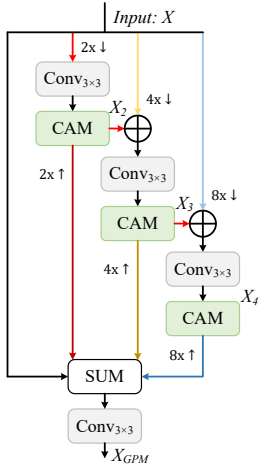

Fig. 7. Detailed structure of the proposed GPM.

In the spatial domain, given the input feature $X \in \mathbb{R}^{H \times W \times C}$, a 1×1 convolution is first applied to increase the channel dimension. The resulting feature is then split into two parts: $X^1$ and $X^2$. The former is used to extract local multi-scale features, while the latter serves as a residual connection for later fusion. The construction of local multi-scale features is as follows:

$$X_S = Concat(DConv_{3\times3}(X^1), DConv_{5\times5}(X^1)), \quad (10)$$

where, $DConv_{3\times3}$ and $DConv_{5\times5}$ represent depthwise convolutions [47] with kernel sizes of $3 \times 3$ and $5 \times 5$, respectively.

In the frequency domain, the output $X_S$ is converted into real components $X_R$ and imaginary components $X_I$ through FFT. These components are then concatenated and passed through a convolution. After modulation, the real component and the imaginary component are separated, and IFFT converts the frequency domain features back to the spatial domain to obtain the feature $X_F$:

$$X_R, X_I = FFT(X_S), \quad (11)$$

$$X_F = IFFT\left(ReLU\left(BN(Conv[X_R, X_I])\right)\right), \quad (12)$$

To complete the fusion, $X_F$ is first reduced in channel dimension and then added element-wise to $X^2$. A CAM (Section III-B2) is finally applied to refine fused feature, enhancing the model's focus on target-relevant information and further improving detection performance.

Fig. 6 presents the visualization results of the proposed MFM in different scenarios. Without MFM, the feature maps exhibit substantial background interference (highlighted in the yellow boxes), which may lead to false alarms. With MFM, target regions are significantly enhanced while background responses are suppressed, resulting in improved target saliency and reduced interference.

### D. Global Pooling Module

While U-Net effectively fuses multi-scale features, its top-down architecture tends to dilute high-level semantics during decoding, thereby limiting the localization of small targets. To address this, the GPM is constructed to extract high-level context and propagate it to each decoder layer via global semantic guidance flows, enhancing the accuracy of target localization.

As illustrated in Fig. 7, the GPM constructs a four-branch hierarchical structure to aggregate global contextual information at multiple resolutions. The input feature map $X \in \mathbb{R}^{H \times W \times C}$ is first processed through multiple average pooling layers with different downsampling rates, generating multi-scale representations. Each downsampled feature map is passed through a $3 \times 3$ convolution followed by a CAM, which dynamically adjusts the importance of each channel based on semantic relevance. To facilitate information exchange across branches, skip connections are introduced between adjacent scales. Specifically, finer-scale feature maps are downsampled and transmitted to coarser branches, where they are fused via element-wise addition before further convolutional processing. This hierarchical interaction enhances cross-scale feature integration and enriches semantic representation. The above process can be expressed as:

$$X_2 = CAM\left(Conv_{3\times3}\left(\mathcal{P}_{avg\ s=2}^{k=2}(X)\right)\right), \quad (13)$$

$$X_3 = CAM\left(Conv_{3\times3}\left(\mathcal{P}_{avg\ s=4}^{k=4}(X) \oplus \mathcal{P}_{avg\ s=2}^{k=2}(X_2)\right)\right), \quad (14)$$

$$X_4 = CAM\left(Conv_{3\times3}\left(\mathcal{P}_{avg\ s=8}^{k=8}(X) \oplus \mathcal{P}_{avg\ s=2}^{k=2}(X_3)\right)\right). \quad (15)$$

where $\mathcal{P}_{avg\ s=b}^{k=a}$ represents the average pooling with kernel size $a$ and stride $b$, and $\oplus$ indicates element-wise addition.

Finally, the outputs from all branches are upsampled to the original resolution, concatenated with the input feature map, and fused through a $3 \times 3$ convolution to generate the final output $X_{GPM}$. This design enhances the network's ability to encode multi-scale context, thereby improving the robustness of small target localization.

### E. Loss Function

To enhance the alignment between the global semantic features generated by the GPM and the ground truth, the Soft-IoU loss [48] is employed to train the model. The total loss of the network is formulated as:

$$\mathcal{L}_{total} = \mathcal{L}_{Soft-IoU}(O_{final}, GT) + \mathcal{L}_{Soft-IoU}(X_{GPM}, GT), \quad (16)$$

where $O_{final}$ refers to the final output of the decoder.

## IV. EXPERIMENTS

### A. Datasets

To comprehensively evaluate the effectiveness and superiority of the proposed FSGNet, we conduct experiments on four widely recognized benchmark datasets: NUAA-SIRST [8], IRSTD-1K [27], NUDT-SIRST [9], and SIRSTAUG [28], comprising 427, 1001, 1327, and 9070 annotated infrared images, respectively. These datasets cover a wide variety of infrared scenes, ensuring a rigorous and representative evaluation environment.

To further assess the robustness of FSGNet under noisy environments, we evaluate it on the NoisySIRST dataset [42], which is derived from NUAA-SIRST and adds Gaussian white noise of different intensity levels ($\sigma_n = 10, 20, 30$) to simulate real-world degradation of infrared imagery.

### B. Implementation Details

In this paper, AdamW [49] is adopted as the optimizer for 600 training epochs, with an initial learning rate of 0.001, and a weight decay of $10^{-2}$. The learning rate is gradually reduced to $1 \times 10^{-5}$ using a cosine annealing strategy to ensure stable convergence. To mitigate overfitting, data augmentation is applied to the training images through random flipping and rotatation. Given the large variation in data distribution and resolution across datasets, hyperparameters are fine-tuned for each dataset individually, as summarized in Table I. And four independent models are trained on each dataset. In the inference phase, model predictions are resized to the original image resolution and evaluated using multiple quantitative metrics. The segmentation threshold of the saliency map is set to 0.5.

### C. Evaluation Metrics

To provide a comprehensive evaluation of detection performance, we employ four widely used metrics: intersection over union (IoU), normalized intersection over union (nIoU) [8], detection probability ($P_d$) and false alarm rate ($F_a$). They are defined as follows:

$$IoU = \frac{TP}{T + P - TP}, \quad (17)$$

$$nIoU = \frac{1}{N}\sum_{i}^{N}\frac{TP(i)}{T(i) + P(i) - TP(i)}, \quad (18)$$

$$P_d = \frac{TP}{TP + FN}, \quad (19)$$

$$F_a = \frac{FP}{FP + TN}, \quad (20)$$

where $N$ is the total number of samples, $T$ and $P$ represent the number of true values and predicted positive pixels, respectively. $TP$, $FP$, $TN$ and $FN$ refer to the number of true positive, false positive, true negative, and false negative pixels. Considering the presence of high background clutter in IRSTD, we additionally employ the receiver operating characteristic (ROC) curve to analyze the trade-off between the true positive rate (TPR) and the false positive rate (FPR), thereby providing a more robust and comprehensive assessment of model performance.

### D. Comparison With SOTA Methods

The proposed FSGNet with sixteen representative IRSTD methods, including eight traditional methods (FKRW [50], TLLCM [51], NWMTH [52], GSWLCM [53], IPI [54], RIPT [6], NOLC [55], PSTNN [56]) and eight deep learning-based methods (ACM [8], RDIAN [57], AGPCNet [28], DNANet [9], UIUNet [10], RPCANet [58], MSHNet [59] and L²SKNet [31]). Notably, both ACM [8] and L²SKNet [31] adopt U-Net as their backbone architecture. To ensure a fair comparison, all deep learning models are retrained using the same training benchmark as FSGNet. The training settings strictly follow the original implementations, with consistent hyperparameters applied across all models to eliminate potential bias.

### E. Quantitative Results

To evaluate the detection performance of FSGNet, sixteen

7TABLE I
CUSTOM HYPER-PARAMETERS FOR FIVE DATASETS

| Dataset | NUAA-SIRST [8] | IRSTD-1K [27] | NUDT-SIRST [9] | SIRSTAUG [28] | NoisySIRST [42] |
|---|---|---|---|---|---|
| Batch | 16 | 4 | 16 | 16 | 16 |
| Resolution | 256×256 | 512×512 | 256×256 | 256×256 | 320×320 |
| Train image | 213 | 800 | 663 | 8525 | 341 |
| Test image | 214 | 201 | 664 | 545 | 86 |

TABLE II
COMPARISON WITH SOTA METHODS ON FOUR DATASETS. IN IoU(%), nIoU(%), $P_d$(%), $F_a$($10^{-6}$). THE BEST RESULTS ARE IN BOLD, AND THE SECOND BEST RESULTS ARE IN UNDERLINE.

| Methods | NUAA-SIRST | | | | IRSTD-1K | | | | NUDT-SIRST | | | | SIRSTAUG | | | |
|---|---|---|---|---|---|---|---|---|---|---|---|---|---|---|---|---|
| | IoU↑ | nIoU↑ | $P_d$↑ | $F_a$↓ | IoU↑ | nIoU↑ | $P_d$↑ | $F_a$↓ | IoU↑ | nIoU↑ | $P_d$↑ | $F_a$↓ | IoU↑ | nIoU↑ | $P_d$↑ | $F_a$↓ |
| *Background Suppression Methods* | | | | | | | | | | | | | | | | |
| FKRW [50] | 22.06 | 28.08 | 81.77 | 16.32 | 10.39 | 16.25 | 69.54 | 24.37 | 12.67 | 21.73 | 79.51 | 67.13 | 10.05 | 16.56 | 72.42 | 66.87 |
| TLLCM [51] | 18.66 | 27.71 | 82.24 | 16.95 | 10.26 | 18.24 | 68.55 | 24.48 | 11.08 | 23.82 | 77.97 | 67.26 | 13.96 | 16.80 | 77.47 | 69.71 |
| NWMTH [52] | 15.77 | 17.28 | 71.12 | 55.61 | 18.94 | 16.91 | 49.26 | 21.72 | 11.72 | 10.26 | 52.71 | 46.81 | 18.71 | 16.74 | 62.71 | 32.46 |
| GSWLCM [53] | 15.42 | 13.24 | 72.68 | 21.73 | 12.82 | 13.72 | 70.24 | 13.92 | 10.82 | 18.71 | 67.32 | 53.31 | 21.62 | 19.85 | 69.41 | 42.71 |
| *Low-rank and Sparse Decomposition Methods* | | | | | | | | | | | | | | | | |
| IPI [54] | 25.67 | 33.57 | 85.55 | 11.47 | 27.92 | 30.12 | 81.37 | 16.18 | 28.63 | 38.18 | 74.49 | 41.23 | 25.16 | 34.64 | 76.26 | 43.41 |
| RIPT [6] | 11.05 | 19.91 | 79.08 | 22.61 | 14.11 | 17.43 | 77.55 | 28.31 | 29.17 | 36.12 | 91.85 | 344.3 | 24.13 | 33.98 | 78.54 | 56.24 |
| NOLC [55] | 26.64 | 35.68 | 81.62 | 17.44 | 12.39 | 22.18 | 75.38 | 21.94 | 23.87 | 34.90 | 85.47 | 58.20 | 12.67 | 20.87 | 74.66 | 67.31 |
| PSTNN [56] | 22.40 | 29.59 | 77.95 | 29.11 | 24.57 | 28.71 | 71.99 | 35.26 | 27.72 | 39.80 | 66.13 | 44.17 | 19.14 | 27.16 | 73.14 | 61.58 |
| *Deep Learning Methods* | | | | | | | | | | | | | | | | |
| ACM [8] | 65.28 | 65.67 | 90.49 | 47.13 | 58.84 | 58.23 | 91.92 | 27.59 | 60.63 | 60.94 | 91.75 | 20.06 | 73.84 | 69.83 | 97.52 | 76.35 |
| RDIAN [57] | 70.66 | 74.04 | 93.92 | 43.08 | 63.37 | 62.63 | 88.55 | 30.18 | 77.77 | 79.46 | 95.66 | 26.36 | 74.19 | 69.80 | **99.17** | 23.97 |
| AGPCNet [28] | 75.69 | 76.60 | <u>96.48</u> | <u>14.99</u> | 66.29 | 65.23 | <u>92.83</u> | 13.12 | 88.87 | 90.64 | 97.20 | 10.02 | 74.71 | <u>71.49</u> | 97.67 | 34.84 |
| DNANet [9] | 76.34 | **79.19** | 95.82 | 15.23 | 66.50 | 66.13 | 91.58 | 18.31 | 91.96 | <u>92.88</u> | <u>98.94</u> | 9.28 | <u>74.88</u> | 70.23 | 97.80 | 30.07 |
| UIUNet [10] | <u>77.17</u> | 78.86 | 95.44 | **14.34** | 65.34 | 65.55 | 90.91 | 15.01 | 90.43 | 89.87 | <u>98.94</u> | 7.72 | 73.82 | 70.36 | 98.49 | <u>19.35</u> |
| RPCANet [58] | 53.93 | 61.01 | 95.44 | 149.2 | 62.09 | 61.48 | 89.35 | 48.61 | 87.79 | 89.80 | 95.98 | 32.03 | 72.60 | 69.55 | 97.66 | 351.5 |
| MSHNet [59] | 73.88 | 72.56 | 95.82 | 19.25 | 67.17 | 60.38 | 92.52 | <u>12.60</u> | 76.95 | 79.72 | 95.77 | 20.52 | 73.06 | 68.87 | 94.22 | 263.1 |
| L$^2$SKNet [31] | 72.90 | 73.89 | 93.92 | 40.34 | <u>67.38</u> | <u>66.36</u> | 91.58 | 14.78 | <u>93.52</u> | 92.86 | 97.57 | <u>5.29</u> | 73.49 | 69.78 | 98.35 | 43.03 |
| FSGNet (ours) | **78.48** | <u>79.09</u> | **96.58** | 21.95 | **72.45** | **68.04** | **92.93** | **5.43** | **93.78** | **94.14** | **99.26** | **4.89** | **75.69** | **71.70** | <u>98.76</u> | **14.36** |

TABLE III
PERFORMANCE COMPARISON OF DEEP LEARNING-BASED METHODS ON THE NOISYSIRST DATASET WITH DIFFERENT GAUSSIAN WHITE NOISES

| Methods | $\sigma_n$ = 10 (SNR=5.35) | | $\sigma_n$ = 20 (SNR=3.69) | | $\sigma_n$ = 30 (SNR=2.76) | |
|---|---|---|---|---|---|---|
| | IoU | nIoU | IoU | nIoU | IoU | nIoU |
| ACM [8] | 67.36 | 68.01 | 68.55 | 64.59 | 62.18 | 62.11 |
| RDIAN [57] | 70.91 | 75.71 | 69.27 | 70.42 | 65.99 | 66.41 |
| AGPCNet [28] | 71.41 | 70.96 | 71.30 | 69.90 | 68.05 | 67.28 |
| DNANet [9] | 77.01 | 74.27 | 70.56 | 68.77 | 68.59 | 62.89 |
| UIUNet [10] | <u>77.77</u> | <u>74.74</u> | <u>74.53</u> | <u>72.27</u> | 68.70 | **70.19** |
| RPCANet [58] | 65.44 | 67.83 | 50.58 | 55.11 | 44.46 | 48.81 |
| MSHNet [59] | 73.54 | 71.21 | 70.91 | 68.40 | <u>68.77</u> | 64.12 |
| L$^2$SKNet [31] | 72.29 | 72.73 | 71.06 | 71.12 | 64.14 | 64.19 |
| FSGNet (ours) | **77.98** | **77.54** | **75.16** | **74.04** | **72.24** | <u>69.96</u> |

SOTA methods are compared on four public datasets: NUAA-SIRST, IRSTD-1K, NUDT-SIRST, and SIRSTAUG based on four widely adopted metrics: IoU, nIoU, $P_d$, and $F_a$. Table II summarizes the comparative results, followed by detailed analysis:

1) FSGNet consistently achieves superior performance across the evaluated datasets and metrics, demonstrating strong generalization and robustness. For example, on the IRSTD-1K dataset, the proposed FSGNet yields improvements of 5.07% in IOU and 1.68% in nIOU over L$^2$SKNet, a relatively advanced baseline, while also reducing the $F_a$ by 9.35%, indicating enhanced detection precision under complex background conditions. On NUAA-SIRST dataset, FSGNet exceeds UIUNet by achieving improvements of 1.31% in IoU and 0.23% in nIoU, while maintaining a comparable $F_a$.

2) Deep learning-based methods significantly outperform traditional approaches across all datasets, especially in terms of IoU and nIoU metrics, with a gain of nearly three times, which highlights the limitations of handcrafted features and prior-driven algorithms in capturing high-level semantic structures. In contrast, learning-based models demonstrate greater adaptability and richer feature abstraction, thereby improving robustness to scene variation.

3) Compared with L$^2$SKNet and UIUNet, FSGNet exhibits consistently better trade-offs between accuracy and false alarm control. On the SIRSTAUG dataset, it achieves an IOU of 75.69%, with an improvement of 2.2% over L$^2$SKNet, while reducing $F_a$ from 43.03% to 14.36%. On NUDT-SIRST dataset, the proposed method achieves the highest $P_d$ of 99.26%, surpassing L$^2$SKNet by 1.69%, and attains the lowest $F_a$ of 4.89%, confirming its reliability in complex environments.

Meanwhile, the experimental results of FSGNet on the noisy dataset NoisySIRST [42] are shown in Table III, which clearly shows that the proposed method nearly outperforms existing SOTA methods at all noise levels in terms of IoU and nIoU evaluation metrics, fully demonstrating its excellent noise resistance. And Fig. 8 also presents the ROC curves of various methods, further demonstrating the superiority of FSGNet. The ROC curve of FSGNet consistently surpasses those of competing approaches, indicating a more optimal balance between TPR and FPR.

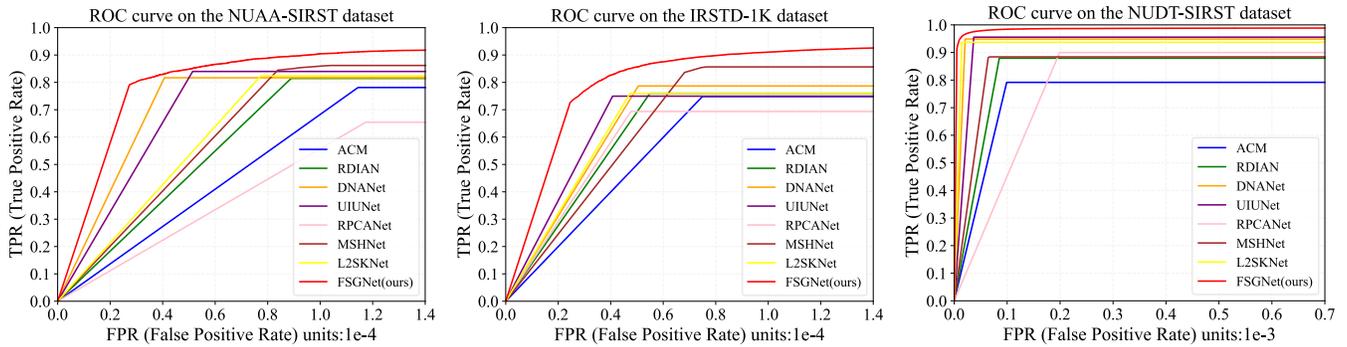

Fig. 8. ROC curves of different IRSTD methods on three public datasets.

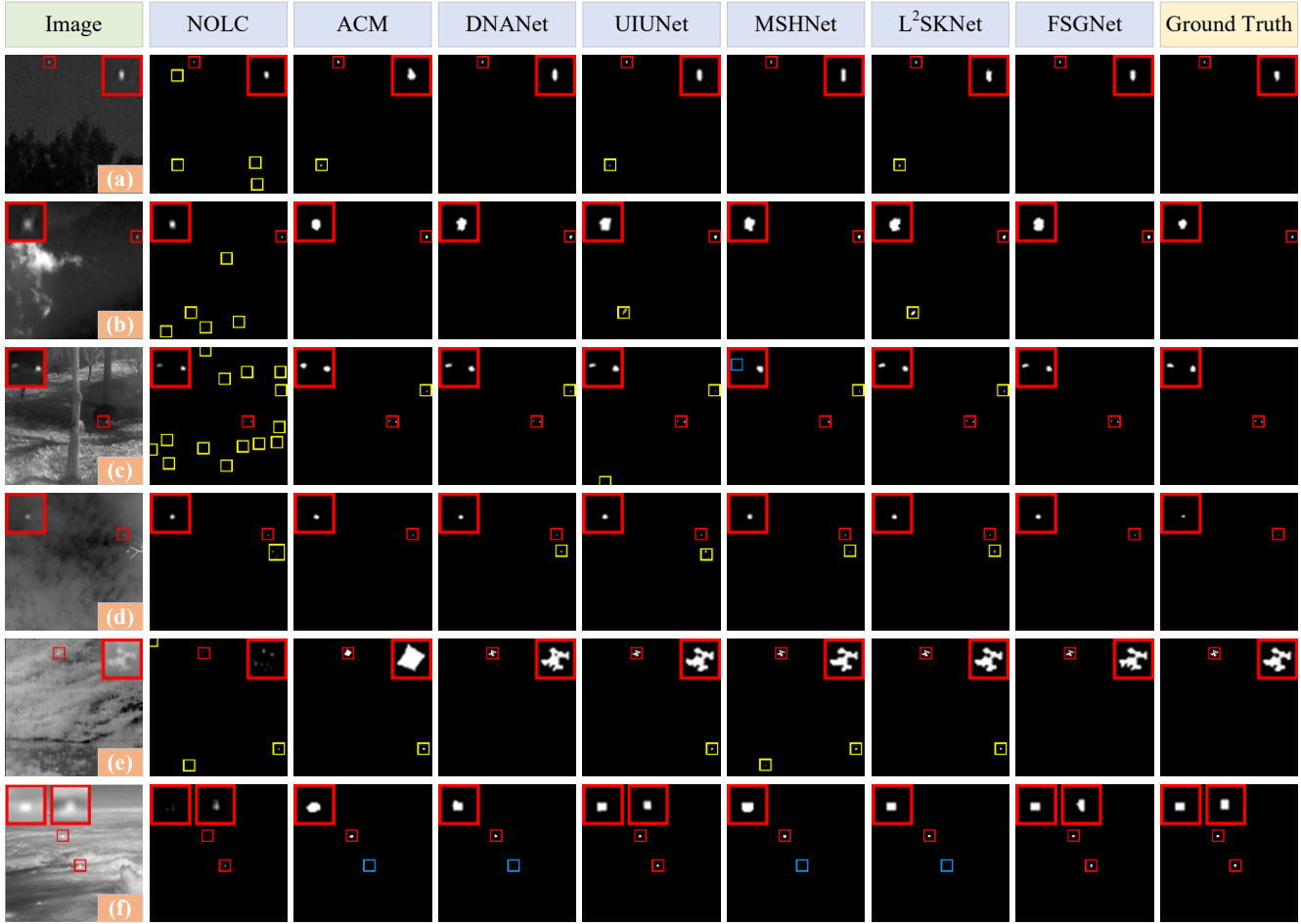

Fig. 9. Visualization of different IRSTD methods on NUAA-SIRST, IRSTD-1K, and NUDT-SIRST datasets. The red, blue, and yellow boxes represent correctly detected targets, miss detection, and false alarms, respectively.

*F. Visual Results*

Fig. 9 shows the qualitative comparison results of various IRSTD methods on three benchmark datasets, including NUAA-SIRST (Fig. 9(a)-(b)), IRSTD-1K (Fig. 9(c)-(d)), and NUDT-SIRST (Fig. 9(e)-(f)). Compared with traditional methods, deep learning-based methods significantly reduce false alarms under most scenarios. Specifically, traditional methods exhibit a high number of false alarms in Fig. 9(a)-(c), and although a target is identified in Fig. 9(f), the blurred contour undermines precise recognition. The strength of deep learning methods lies in their capabilities to autonomously extract features, thereby enhancing robustness against background interference. While other SOTA methods demonstrate competitive detection performance, they generally underperform compared to our method. As shown in Fig. 9(f), all methods except for UIUNet and our model suffer from missed detections; and in other scenarios, UIUNet produces notable false alarms. In contrast, the proposed method consistently achieves high detection accuracy and robustness across diverse scenarios, confirming its superiority.

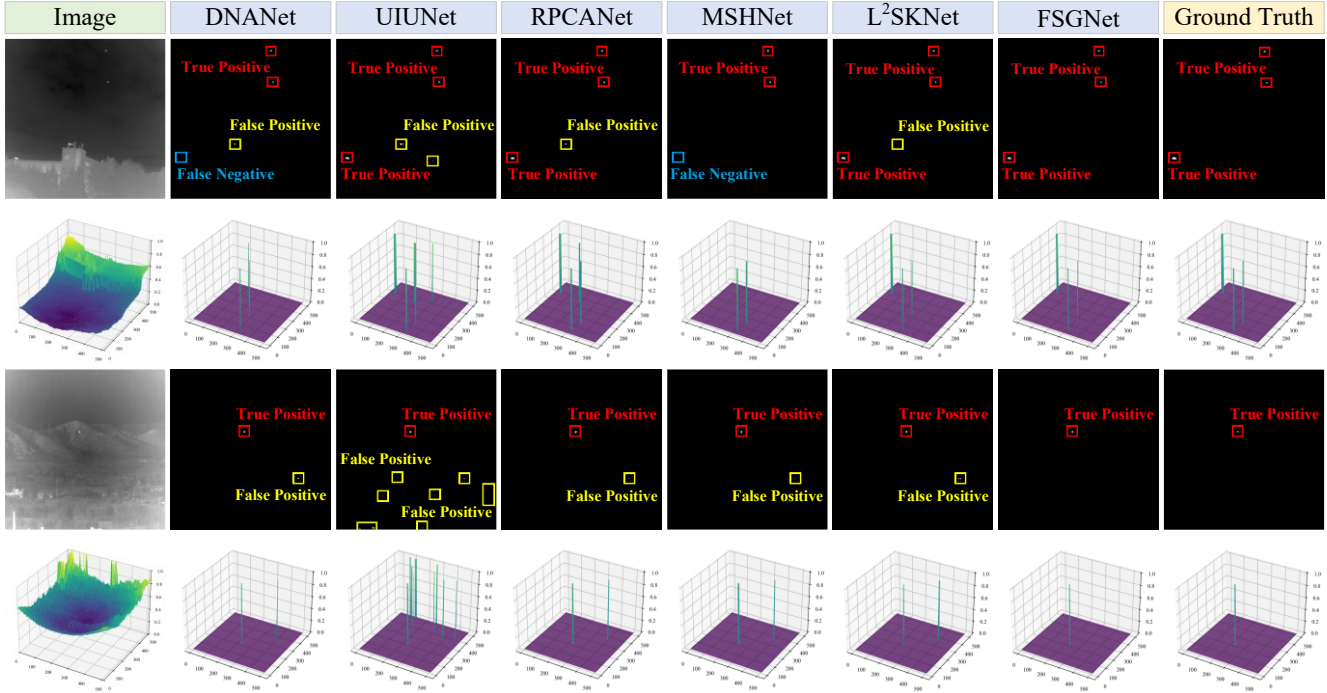

Fig. 10. Visualization of different IRSTD methods on IRSTD-1K dataset. The red, blue, and yellow boxes represent correctly detected targets, miss detection, and false alarms, respectively.

TABLE IV
ABLATION STUDY TO EVALUATE THE MIAM, MFM, GPM AND GSGM MODULE, IN IoU(%) AND nIoU (%) ON FOUR DATASETS.

| Strategy | MIAM | MFM | GPM | GSGM | NUAA-SIRST IoU↑ | nIoU↑ | IRSTD-1K IoU↑ | nIoU↑ | NUDT-SIRST IoU↑ | nIoU↑ | SIRSTAUG IoU↑ | nIoU↑ |
|---|---|---|---|---|---|---|---|---|---|---|---|---|
| (a) | × | × | × | × | 74.99 | 73.28 | 64.75 | 63.24 | 85.42 | 85.58 | 69.57 | 66.12 |
| (b) | √ | × | × | × | 76.15 | 78.24 | 67.18 | 68.71 | 91.85 | 92.01 | 74.14 | 71.30 |
| (c) | √ | √ | × | × | 76.76 | 78.13 | 68.36 | **68.89** | 93.18 | 93.45 | 74.91 | 71.92 |
| (d) | √ | √ | √ | × | 76.83 | 77.90 | 69.21 | 67.97 | 93.69 | 93.85 | 75.41 | **72.09** |
| (e) | √ | √ | √ | √ | **78.48** | **79.09** | **72.45** | 68.04 | **93.78** | **94.14** | **75.69** | 71.70 |

Moreover, the 3D visualization results in Fig. 10 further illustrate the performance of deep learning-based methods. Two representative infrared images from the IRSTD-1K dataset are visualized, revealing that in the presence of dense targets and complex backgrounds, competing methods often fail to detect targets reliably. Notably, UIUNet produces numerous false alarms when targets are faint, as seen in Fig. 10(b). These observations further substantiate the effectiveness and robustness of our method in complex environments.

### G. Ablation Study

#### 1) Performance Analysis among Each Modules

To evaluate the effectiveness of each proposed component, we adopt U-Net [7] as the baseline and incrementally integrate MIAM, MFM, GPM, and GSGM. As shown in Table IV, each module contributes independently to performance improvement. Incorporating MIAM leads to significant IoU gains, with increases of 4.57% on SIRSTAUG dataset and 6.43% on NUDT-SIRST dataset. Adding MFM further improves detection accuracy by effectively reducing false alarms through frequency-domain filtering. The integration of GPM enhances semantic representation, yielding an additional 1.43% IoU gain on SIRSTAUG dataset. Finally, the inclusion of GSGM substantially improves localization accuracy by reinforcing high-level semantic guidance, confirming its critical role in the decoder.

Fig. 11 presents visualization results to illustrate the impact of progressively integrating individual modules on IRSTD performance. The results show that the U-Net baseline suffers from missed detections and false alarms, whereas the progressive integration of each module enhances target response and suppresses background interference. Ultimately, the proposed FSGNet achieves the most accurate detection results, validating both the individual effectiveness and the complementary benefits of each module.

#### 2) Performance Analysis of MIAM Internal Components

The proposed MIAM enhances feature representation by integrating PConv into the residual structure and incorporating attention mechanisms. As shown in Table V, the inclusion of PConv improves the baseline IoU from 70.75% to 71.21% on the IRSTD-1K dataset, though false alarms remain due to background clutter. Incorporating both channel and spatial attention mechanisms further increases the IoU to 72.45% and nIoU to 68.04%, indicating improved target discrimination and background suppression. On the NUAA-SIRST dataset, MIAM achieves the highest IoU of 78.48% and nIoU of 79.09%, surpassing the 77.35% and 78.51% results obtained without attention, highlighting the effectiveness of joint attention mechanisms in focusing on salient target regions. These results confirm that the combination of receptive field expansion and attention mechanisms enhances detection accuracy while suppressing false positives, contributing to robust performance across diverse infrared scenarios.

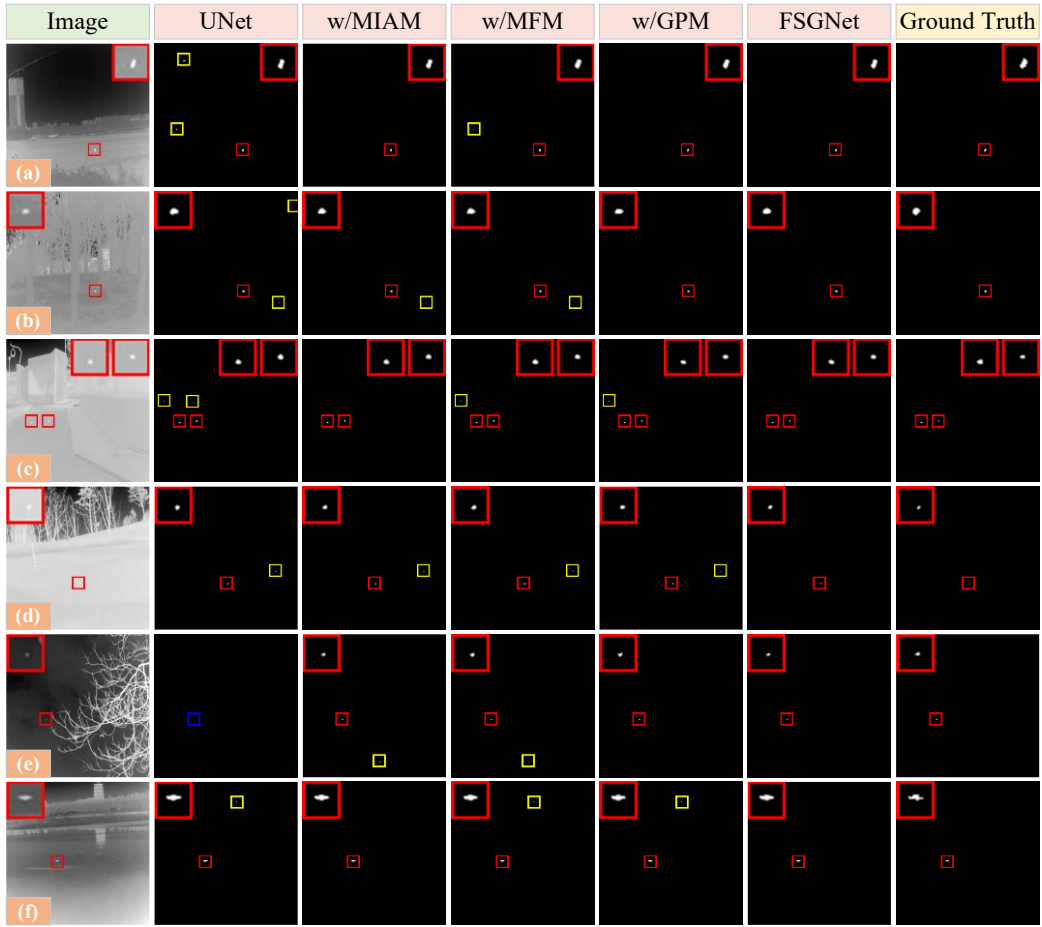

Fig. 11. Visualization of different IRSTD methods on NUAA-SIRST, IRSTD-1K, and NUDT-SIRST datasets. From left to right, they are the original image, U-Net baseline, w/MIAM, w/MFM, w/GPM, FSGNet and Ground Truth. The red, blue, and yellow boxes represent correctly detected targets, miss detection, and false alarms, respectively.

TABLE V
ABLATION STUDY TO EVALUATE THE COMPONENTS OF MIAM, IN IoU(%) AND nIoU (%) ON FOUR DATASETS.

| Strategy | Module | | | | NUAA-SIRST | | IRSTD-1K | | NUDT-SIRST | | SIRSTAUG | |
|---|---|---|---|---|---|---|---|---|---|---|---|---|
| | PConv | Residual Connections | CAM | SAM | IoU↑ | nIoU↑ | IoU↑ | nIoU↑ | IoU↑ | nIoU↑ | IoU↑ | nIoU↑ |
| (a) | √ | × | × | × | 73.77 | 76.18 | 70.75 | 64.93 | 93.15 | 93.47 | 72.53 | 71.19 |
| (b) | √ | √ | × | × | 77.35 | 78.51 | 71.21 | 66.92 | 93.25 | 93.67 | 75.17 | 71.32 |
| (c) | √ | √ | √ | × | 77.50 | 78.16 | 71.92 | **68.79** | 93.28 | 93.53 | 75.35 | 71.68 |
| (d) | √ | √ | √ | √ | **78.48** | **79.09** | **72.45** | 68.04 | **93.78** | **94.14** | **75.69** | **71.70** |

*3) Performance Analysis of MFM Internal Components*

The proposed MFM integrates spatial-domain feature extraction with frequency-domain fusion to refine skip connections. As shown in Table VI, the progressive addition of MFM components consistently improves detection performance. Compared with different-scale convolution integration, the incorporation of CAM and FFT yields greater performance gains. For instance, CAM improves IoU by 2.3% on the IRSTD-1K dataset. Moreover, across all four datasets, FFT integration produces superior results across all evaluation metrics, further validating its effectiveness. In summary, the proposed MFM effectively suppresses false alarms caused by background clutter and enhances the robustness of target detection.

TABLE VI
ABLATION STUDY TO EVALUATE THE COMPONENTS OF MFM, IN IoU(%) AND nIoU (%) ON FOUR DATASETS.

| Strategy | Module | | | | NUAA-SIRST | | IRSTD-1K | | NUDT-SIRST | | SIRSTAUG | |
|---|---|---|---|---|---|---|---|---|---|---|---|---|
| | Dconv 3×3 | Dconv 5×5 | CAM | FFT | IoU↑ | nIoU↑ | IoU↑ | nIoU↑ | IoU↑ | nIoU↑ | IoU↑ | nIoU↑ |
| (a) | √ | × | × | × | 76.20 | 76.97 | 67.52 | **68.99** | 93.47 | 93.51 | 73.20 | 69.76 |
| (b) | √ | √ | × | × | 76.33 | 77.93 | 70.11 | 65.80 | 93.51 | 93.69 | 74.68 | 71.12 |
| (c) | √ | √ | √ | × | 77.21 | 77.90 | 72.41 | 67.45 | 93.55 | 93.83 | 74.78 | 70.60 |
| (d) | √ | √ | √ | √ | **78.48** | **79.09** | **72.45** | <u>68.04</u> | **93.78** | **94.14** | **75.69** | **71.70** |

TABLE VII
ABLATION STUDY TO EVALUATE THE COMPONENTS OF GSGM, IN IoU(%) AND nIoU (%) ON FOUR DATASETS.

| Strategy | The number of GSGFs | | | | NUAA-SIRST | | IRSTD-1K | | NUDT-SIRST | | SIRSTAUG | |
|---|---|---|---|---|---|---|---|---|---|---|---|---|
| | 2×↑ | 4×↑ | 8×↑ | 16×↑ | IoU↑ | nIoU↑ | IoU↑ | nIoU↑ | IoU↑ | nIoU↑ | IoU↑ | nIoU↑ |
| (a) | √ | × | × | × | 76.83 | 77.90 | 69.21 | 67.97 | 93.69 | 93.85 | 75.41 | **72.09** |
| (b) | √ | √ | × | × | 77.39 | 78.18 | 69.76 | 66.89 | 93.60 | 93.99 | 75.29 | 71.19 |
| (c) | √ | √ | √ | × | 77.86 | 78.68 | 71.34 | 68.14 | 93.54 | 93.58 | 75.50 | 71.47 |
| (d) | √ | √ | √ | √ | **78.48** | **79.09** | **72.45** | **68.04** | **93.78** | **94.14** | **75.69** | <u>71.70</u> |

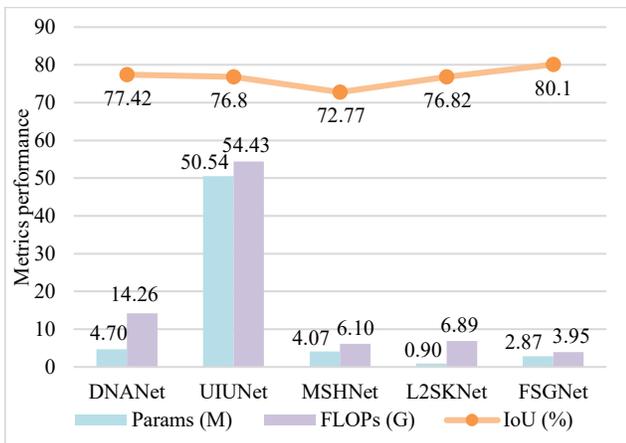

Fig. 12. Performance evaluation of five algorithms on three metrics.

*4) Performance Analysis of GSGM Internal Components*

To explore the effect of different numbers of GSGFs on propagating high-level semantic information to each decoder layer, we conducted an ablation study, as presented in Table VII. The results indicate that increasing the number of semantic guidance flows consistently enhances detection performance. When four semantic guidance flows are employed, the model achieves its optimal performance. This demonstrates that the high-level semantic information generated by GPM, guided by the global semantic flows, effectively compensates for the lack of semantic representation in the decoder, thereby significantly enhancing the localization capability for small targets. These findings validate the effectiveness and necessity of the proposed global semantic guidance mechanism in improving the accuracy of small target detection.

*H. Model Complexity Analysis*

This section presents a comparative analysis of model complexity among recent deep learning-based IRSTD methods by evaluating their average IoU scores on four benchmark datasets, along with their corresponding model parameters (Params) and floating-point operations (FLOPs), as illustrated in Fig. 12. The results indicate that the proposed FSGNet achieves the highest average IoU, outperforming the second-best method by 2.68%. Moreover, FSGNet exhibits strong model efficiency, attaining the lowest FLOPs while ranking second in the number of model parameters. Despite delivering superior detection performance, FSGNet significantly reduces computational overhead, thereby demonstrating an effective trade-off between accuracy and efficiency.

## V. CONCLUSION

To address the challenge that deep semantic information is gradually diluted during the transmission process in U-Net architectures for IRSTD, which hinders the precise localization of small targets, this paper proposes a lightweight and effective network, termed FSGNet. Considering the small size and low contrast of infrared small targets, we first design a multi-directional interactive attention module by integrating a pinwheel-shaped convolution and the convolutional block attention module, thereby enhancing the network's focus and robustness on target regions. Then, a multi-scale frequency-aware module is proposed to address the issue that complex background noise is easily transmitted from the encoder to the decoder through skip connections, leading to false alarms. At the bottom of the encoder, a global pooling module is constructed to capture high-level semantic features across the entire image effectively. Additionally, global semantic guidance flows are utilized to propagate localization information to each decoder layer in a semantically guided manner. Despite its relatively simple architecture, FSGNet establishes new performance benchmarks on multiple public datasets. It significantly improves detection performance while maintaining low parameters and computational complexity, demonstrating its efficiency and practicality for the IRSTD task.
## REFERENCES

[1] F. Lin, K. Bao, Y. Li, D. Zeng and S. Ge, "Learning Contrast-

Enhanced Shape-Biased Representations for Infrared Small Target Detection," *IEEE Trans. Image Process.*, vol. 33, pp. 3047-3058, 2024.

[2] F. Liu, C. Gao, F. Chen, D. Meng, W. Zuo and X. Gao, "Infrared Small and Dim Target Detection with Transformer Under Complex Backgrounds," *IEEE Trans. Image Process.*, vol. 32, pp. 5921-5932, 2023.

[3] B. Yang, S. Zhao, W. Wang, J. Luo, H. Pu, M. Zhou, and Y. Pi, "MTMLNet: Multi-task Mutual Learning Network for Infrared Small Target Detection and Segmentation," *IEEE Trans. Image Process.*, 2025, doi: 10.1109/TIP.2025.3587576.

[4] J.-F. Rivest and R. Fortin, "Detection of dim targets in digital infrared imagery by morphological image processing," *Opt. Eng.*, vol. 35, no. 7, pp. 1886-1893, 1996.

[5] J. Han et al., "Infrared Small Target Detection Based on the Weighted Strengthened Local Contrast Measure," *IEEE Geosci. Remote Sens. Lett.*, vol. 18, no. 9, pp. 1670-1674, 2021.

[6] Y. Dai and Y. Wu, "Reweighted Infrared Patch-Tensor Model with Both Nonlocal and Local Priors for Single-Frame Small Target Detection," *IEEE J. Sel. Topics Appl. Earth Observ. Remote Sens.*, vol. 10, no. 8, pp. 3752-3767, 2017.

[7] O. Ronneberger, P. Fischer, and T. Brox, "U-Net: Convolutional Networks for Biomedical Image Segmentation," in *Proc. 18th Int. Conf. Med. Image Comput. Comput.-Assist. Intervent.*, Munich, Germany. Cham, Switzerland: Springer, pp. 234-241, 2015.

[8] Y. Dai, Y. Wu, F. Zhou, and K. Barnard, "Asymmetric Contextual Modulation for Infrared Small Target Detection," in *Proc. IEEE/CVF Winter Conf. Appl. Comput. Vis (WACV).*, pp. 950-959, 2021.

[9] B. Li, C. Xiao, L. Wang, Y. Wang, Z. Lin, M. Li, W. An, and Y. Guo, "Dense Nested Attention Network for Infrared Small Target Detection," *IEEE Trans. Image Process.*, vol. 32, pp. 1745-1758, 2023.

[10] X. Wu, D. Hong, and J. Chanussot, "UIUNet: U-Net in U-Net for Infrared Small Target Detection," *IEEE Trans. Image Process.*, vol. 32, pp. 364-376, 2023.

[11] S. Yuan, H. Qin, X. Yan, N. Akhtar, and A. Mian, "SCTransNet: Spatial-Channel Cross Transformer Network for Infrared Small Target Detection," *IEEE Trans. Geosci. Remote Sens.*, vol. 62, pp. 1-15, 2024.

[12] J.-J. Liu, Q. Hou, M.-M. Cheng, J. Feng, and J. Jiang, "A Simple Pooling-Based Design for Real-Time Salient Object Detection," in *Proc. IEEE/CVF Conf. Comput. Vis. Pattern Recognit. (CVPR)*, pp. 3917-3927, 2019.

[13] K. He, X. Zhang, S. Ren, and J. Sun, "Deep Residual Learning for Image Recognition," in *Proc. IEEE/CVF Conf. Comput. Vis. Pattern Recognit. (CVPR)*, pp. 770-778, 2016.

[14] Yang J, Liu S, Wu J, Su X, Hai N, Huang X, "Pinwheel-shaped Convolution and Scale-based Dynamic Loss for Infrared Small Target Detection," in *Proc. AAAI Conf. Artif. Intell. (AAAI)*, vol. 39, no. 9, pp. 9202-9210, 2025.

[15] S. Woo, J. Park, J. Lee, and I. Kweon, "CBAM: Convolutional Block Attention Module," in *Proc. Eur. Conf. Comput. Vis. (ECCV)*, pp. 3-19, 2018.

[16] Z. Lu, Z. Huang, Q. Song, K. Bai, and Z. Li, "An Enhanced Image Patch Tensor Decomposition for Infrared Small Target Detection," *Remote Sens.*, vol. 14, no. 23, Art. no. 6044, 2022.

[17] L. Deng, G. Xu, J. Zhang, and H. Zhu, "Entropy-Driven Morphological Top-hat Transformation for Infrared Small Target Detection," *IEEE Trans. Aerosp. Electron. Syst.*, vol. 58, no. 2, pp. 962-975, 2022.

[18] L. Deng, J. Zhang, G. Xu, and H. Zhu, "Infrared small target detection via adaptive M-estimator ring top-hat transformation," *Pattern Recognit.*, vol. 112, Art. no. 107729, 2021.

[19] H. Li, Q. Wang, H. Wang, and W. Yang, "Infrared small target detection using tensor based least mean square," *Comput. Electr. Eng.*, vol. 91, Art. no. 106994, 2021.

[20] C. L. P. Chen, H. Li, Y. Wei, T. Xia, and Y. Y. Tang, "A Local Contrast Method for Small Infrared Target Detection," *IEEE Trans. Geosci. Remote Sens.*, vol. 52, no. 1, pp. 574-581, 2013.

[21] Y. Wei, X. You, and H. Li, "Multiscale patch-based contrast measure for small Infrared Small Target Detection," *Pattern Recognit.*, vol. 58, pp. 216-226, 2016.

[22] Y. He, M. Li, J. Zhang, and Q. An, "Small infrared target detection based on low-rank and sparse representation," *Infr. Phys. Technol.*, vol. 68, pp. 98-109, 2015.

[23] H. Zhu, H. Ni, S. Liu, G. Xu, and L. Deng, "TNLRS: Target-Aware Non-Local Low-Rank Modeling With Saliency Filtering Regularization for Infrared Small Target Detection," *IEEE Trans. Image Process.*, vol. 29, pp. 9546-9558, 2020.

[24] T. Zhang, Z. Peng, H. Wu, Y. He, C. Li, and C. Yang, "Infrared small target detection via self-regularized weighted sparse model," *Neurocomputing*, vol. 420, pp. 124-148, 2021.

[25] H. Wang, L. Zhou, and L. Wang, "Miss Detection vs. False Alarm: Adversarial Learning for Small Object Segmentation in Infrared Images," in *Proc. IEEE/CVF Int. Conf. Comput. Vis. (ICCV)*, pp. 8509-8518, 2019.

[26] Y. Dai, Y. Wu, F. Zhou, and K. Barnard, "Attentional Local Contrast Networks for Infrared Small Target Detection," *IEEE Trans. Geosci. Remote Sens.*, vol. 59, no. 11, pp. 9813-9824, 2021.

[27] M. Zhang, R. Zhang, Y. Yang, H. Bai, J. Zhang, and J. Guo, "ISNet: Shape Matters for Infrared Small Target Detection," in *Proc. IEEE/CVF Conf. Comput. Vis. Pattern Recognit. (CVPR)*, pp. 877-886, 2022.

[28] T. Zhang, L. Li, S. Cao, T. Pu, and Z. Peng, "Attention-Guided Pyramid Context Networks for Detecting Infrared Small Target Under Complex Background," *IEEE Trans. Aerosp. Electron. Syst.*, vol. 59, no. 4, pp. 4250-4261, 2023.

[29] M. Xiao et al., "Background Semantics Matter: Cross-Task Feature Exchange Network for Clustered Infrared Small Target Detection with Sky-Annotated Dataset," *arXiv:2407.20078*, 2024.

[30] M. Zhang, Y. Wang, J. Guo, Y. Li, X. Gao, and J. Zhang, "IRSAM: Advancing Segment Anything Model for Infrared Small Target Detection," in *Proc. Eur. Conf. Comput. Vis. (ECCV)*, pp. 233-249, 2024.

[31] F. Wu, A. Liu, T. Zhang, L. Zhang, J. Luo, and Z. Peng, "Saliency at the Helm: Steering Infrared Small Target Detection with Learnable Kernels," *IEEE Trans. Geosci. Remote Sens.*, vol. 63, pp. 1-14, 2025.

[32] A. Gu and T. Dao, "Mamba: Linear-time sequence modeling with selective state spaces," 2023, *arXiv:2312.00752*.

[33] T. Chen, Z. Ye, Z. Tan, T. Gong, Y. Wu, Q. Chu, B. Liu, N. Yu, and J. Ye, "MiM-ISTD: Mamba-in-Mamba for Efficient Infrared Small-Target Detection," *IEEE Trans. Geosci. Remote Sens.*, vol. 62, pp. 1-13, 2024.

[34] M. Zhang, X. Li, F. Gao, and J. Guo, "IRMamba: Pixel Difference Mamba with Layer Restoration for Infrared Small Target Detection", in *Proc. AAAI Conf. Artif. Intell. (AAAI)*, vol. 39, no. 9, pp. 10003-10011, 2025.

[35] Y. Rao, W. Zhao, Z. Zhu, J. Lu, and J. Zhou, "Global Filter Networks for Image Classification," in *Adv. Neural Inf. Process. Syst. (NIPS)*, vol. 34, pp. 980-993, 2021.

[36] Z. Huang, Z. Zhang, C. Lan, Z.-J. Zha, Y. Lu, and B. Guo, "Adaptive Frequency Filters as Efficient Global Token Mixers," in *Proc. IEEE/CVF Int. Conf. Comput. Vis. (ICCV)*, pp. 6049-6059, 2023.

[37] Y. Tatsunami and M. Taki, "FFT-based Dynamic Token Mixer for Vision," in *Proc. AAAI Conf. Artif. Intell. (AAAI)*, vol. 38, no. 14, pp. 15328-15336, 2024.

[38] Y. Huang, X. Zhi, J. Hu, L. Yu, Q. Han, W. Chen, and W. Zhang, "FDDBA-Net: Frequency Domain Decoupling Bidirectional Interactive Attention Network for Infrared Small Target Detection," *IEEE Trans. Geosci. Remote Sens.*, vol. 62, pp. 1-16, 2024.

[39] Y. Zhu, Y. Ma, F. Fan, J. Huang, Y. Yao, X. Zhou, and R. Huang, "Towards Robust Infrared Small Target Detection via Frequency and Spatial Feature Fusion," *IEEE Trans. Geosci. Remote Sens.*, vol. 63, pp. 1-15, 2025.

[40] T. Ma, G. Guo, Z. Li, and Z. Yang, "Infrared Small Target Detection Method Based on High-low Frequency Semantic Reconstruction," *IEEE Geosci. Remote Sens. Lett.*, vol. 21, pp. 1-5, 2024.

[41] M. Xu, C. Yu, Z. Li, H. Tang, Y. Hu and L. Nie, "HDNet: A Hybrid Domain Network with Multiscale High-Frequency Information Enhancement for Infrared Small-Target Detection," *IEEE Trans. Geosci. Remote Sens.*, vol. 63, pp. 1-15, 2025.

[42] Y. Dai, P. Pan, Y. Qian, Y. Li, X. Li, and J. Yang, "Pick of the Bunch: Detecting Infrared Small Targets Beyond Hit-Miss Trade-Offs via Selective Rank-Aware Attention," *IEEE Trans. Geosci. Remote Sens.*, vol. 62, pp. 1-15, Sep. 2024.

[43] Y. Zhang et al., "APTNet: Adaptive Partial Transformer Network for Infrared Small Target Detection," in *IEEE Sens. J.*, vol. 25, no. 10, pp. 17960-17974, 15 May15, 2025

[44] Y. Zhang, W. Bao, Y. Yang, W. Wan, Q. Xiao and X. Zou, "HAFNet: Hierarchical Attention Fusion Network for Infrared Small Target Detection," in *IEEE Trans. Geosci. Remote Sens.*, vol. 63, pp. 1-16, 2025

[45] Y. Zhang, W. Bao, Y. Yang, W. Wan, Q. Xiao and X. Zou, "MPCNet:

Multiscale Perception and Cross-Attention Feature Fusion Network for Infrared Small Target Detection," in *IEEE Trans. Geosci. Remote Sens.*, vol. 64, pp. 1-15, 2026.

[46] Y. Zhong, B. Li, L. Tang, S. Kuang, S. Wu, and S. Ding, "Detecting Camouflaged Object in Frequency Domain," in *Proc. IEEE/CVF Conf. Comput. Vis. Pattern Recognit. (CVPR)*, pp. 4504-4513, 2022.

[47] F. Chollet, "Xception: Deep Learning with Depthwise Separable Convolutions," in *Proc. IEEE/CVF Conf. Comput. Vis. Pattern Recognit. (CVPR)*, pp. 1800-1807, 2017.

[48] M. Rahman and Y. Wang, "Optimizing Intersection-over-union in Deep Neural Networks for Image Segmentation," in *Proc. 12th Int. Symp. Vis. Comput.*, pp. 234-244, 2016.

[49] I. Loshchilov and F. Hutter, "Decoupled Weight Decay Regularization," in *Proc. Int. Conf. Learn. Representations (ICLR)*, New Orleans, LA, USA, 2019.

[50] Y. Qin, L. Bruzzone, C. Gao, and B. Li, "Infrared Small Target Detection Based on Facet Kernel and Random Walker," *IEEE Trans. Geosci. Remote Sens.*, vol. 57, no. 9, pp. 7104-7118, 2019.

[51] J. Han, S. Moradi, I. Faramarzi, C. Liu, H. Zhang, and Q. Zhao, "A Local Contrast Method for Infrared Small-Target Detection Utilizing a Tri-Layer Window," *IEEE Geosci. Remote Sens. Lett.*, vol. 17, no. 10, pp. 1822-1826, 2020.

[52] X. Bai and F. Zhou, "Analysis of new top-hat transformation and the application for infrared dim small target detection," *Pattern Recognit.*, vol. 43, no. 6, pp. 2145-2156, 2010.

[53] Z. Qiu, Y. Ma, F. Fan, J. Huang, and L. Wu, "Global Sparsity-Weighted Local Contrast Measure for Infrared Small Target Detection," *IEEE Geosci. Remote Sens. Lett.*, vol. 19, pp. 1-5, 2022.

[54] C. Gao, D. Meng, Y. Yang, Y. Wang, X. Zhou, and A. G. Hauptmann, "Infrared Patch-Image Model for Small Target Detection in a Single Image," *IEEE Trans. Image Process.*, vol. 22, no. 12, pp. 4996-5009, 2013.

[55] T. Zhang, H. Wu, Y. Liu, L. Peng, C. Yang, and Z. Peng, "Infrared Small Target Detection Based on Non-Convex Optimization with Lp-Norm Constraint," *Remote Sens.*, vol. 11, no. 5, p. 559, 2019.

[56] L. Zhang and Z. Peng, "Infrared Small Target Detection Based on Partial Sum of the Tensor Nuclear Norm," *Remote Sens.*, vol. 11, no. 4, p. 382, 2019.

[57] H. Sun, J. Bai, F. Yang, and X. Bai, "Receptive-Field and Direction Induced Attention Network for Infrared Dim Small Target Detection with a Large-Scale Dataset IRDST," *IEEE Trans. Geosci. Remote Sens.*, vol. 61, pp. 1-13, 2023.

[58] F. Wu, T. Zhang, L. Li, Y. Huang, and Z. Peng, "RPCANet: Deep Unfolding RPCA Based Infrared Small Target Detection," in *Proc. IEEE/CVF Winter Conf. Appl. Comput. Vis. (WACV)*, pp. 4809-4818, 2024.

[59] Q. Liu, R. Liu, B. Zheng, H. Wang, and Y. Fu, "Infrared Small Target Detection with Scale and Location Sensitivity," in *Proc. IEEE/CVF Conf. Comput. Vis. Pattern Recognit. (CVPR)*, pp. 17490-17499, 2024.